\documentclass{article}
\usepackage{ijcai25}

\usepackage{times}
\usepackage{soul}
\usepackage{url}
\usepackage[hidelinks]{hyperref}
\usepackage[utf8]{inputenc}
\usepackage[small]{caption}
\usepackage{graphicx}
\usepackage{amsmath}
\usepackage{amsthm}
\usepackage{booktabs}
\usepackage{algorithm}
\usepackage{algorithmic}
\usepackage[switch]{lineno}
\usepackage{amssymb}

\linenumbers

\title{Leveraging Evolutionary Surrogate-Assisted Prescription in Multi-Objective Chlorination Control Systems}

\author{
    Author Name
    \affiliations
    Affiliation
    \emails
    email@example.com
}

\author{
Rivaaj Monsia\and
Olivier Francon\and
Daniel Young\and
Risto Miikkulainen\\
\affiliations
Project Resilience\\
\emails
rivaaj@utexas.edu,
olivier.francon@cognizant.com,
danyoung@utexas.edu,
risto@cs.utexas.edu
}

\begin{document}
\nolinenumbers

\maketitle


\section{Abstract}
This short, written report introduces the idea of Evolutionary Surrogate-Assisted Prescription (ESP) and presents
preliminary results on its potential use in training real-world agents as a part of the 1st AI for Drinking Water Chlorination Challenge at IJCAI-2025. This work was done by a team from Project Resilience, an organization interested in bridging AI to real-world problems.

\subsection{Background}
The need for reinforcement learning algorithms has exploded as ideas for new agents and agentic systems with real-world applications have been realized. In addition, new perspectives on how to ascertain ideal agents has also expanded-- namely through an evolutionary lens. Leveraging evolutionary principles in commonly multi-objective, noisy fitness landscapes has shown to be advantageous in producing diverse, creative solutions compared to standard RL algorithms. Thus, for the competition at hand, we focus on utilizing the ESP framework ~\cite{francon:esp} to successively train surrogates to model the reward landscape of chlorination control systems, evolve NEAT (Neuroevolution of Augmenting Topologies) ~\cite{stanley:ec02} networks via the surrogate, and collect new data by evaluating the best evolved agents on the RL environment. 

\subsection{Motivation}
This project was undertaken by a group from Project Resilience, an organization which was initiated under the Global Initiative on AI and Data Commons (United Nations ITU) to build and promote a public utility to enable researchers, decision-makers, and domain experts to help contribute to AI initiatives focused on real-world problems. We are especially interested in evolutionary AI and have adapted much previous work, a lot undertaken by members of the team, to build up an evolution-focused RL approach for the competition. 

\section{Methods}
In this section, we briefly present the foundation of the main idea utilized by our approach, ESP, and our specific implementation details including evolutionary configurations, the architecture of agents, etc.

\subsection{Evolutionary Surrogate-Assisted Prescription}

The ESP framework is comprised of two models, a prescriptor ($P_s$) which prescribes actions ($A$) from some observations ($O$) \eqref{eq:1} and a predictor ($P_d$) which models a reward landscape and predicts a reward ($R$) from an observation-action ($O, A$) pair \eqref{eq:2}:
\begin{align}
    P_s(O) = A \label{eq:1}\\
    P_d(C, A) = R \label{eq:2}
\end{align}%

In turn, instead of utilizing an explicit reward function, which may be costly in terms of efficiency, a surrogate/predictor may be utilized, speeding up learning. In addition, Francon \textit{et. al.} showed that ESP converges faster than standard RL methods (PPO, DQN) with a much lower variance and cost/regret. 

ESP takes form as a cycle until agents/prescriptors reach convergence: 

\begin{enumerate}
    \item Collect initial data from a variety of sources/models.
    \item Train a surrogate on said data via gradient descent.
    \item Evolve prescriptors/agents, using the surrogate instead of the explicit reward function.
    \item Apply the best prescriptor(s) on the environment and collect new data
    \item Repeat until convergence of prescriptors.
\end{enumerate}

Although this is a general framework to ESP, it may be, and likely requires to be, expounded upon.

\subsection{Implementation Details}

To collect initial data, we evolved NEAT networks and collected ($C, A, R$) triplets that may be used to train the initial surrogate. In addition, we utilized NSGA-II, a popular multi-objective evolutionary optimization algorithm based on the multi-objective nature of the chlorination systems for the reproduction scheme and fitness evaluation of the NEAT genomes.

The surrogate model was an LSTM-based architecture. 2 separate LSTM's encoded the actions and observations whose outputs were then joined and passed to fully connected layers (with dropout = 0.2, activation function = ReLU). The final output did not have an activation function. The LSTM was provided a sequence of data from 10 consecutive timesteps. At each iteration, the LSTM was trained for 50 epochs (with early stopping) via the Adam optimization algorithm. 

The general configuration for the evolution of the NEAT networks is present in the \texttt{neat-nsga2-config.init} file in the submission. In summary, we evolved a population of 100 genomes with topology mutation rates which were not too extreme. In addition, initial parameters are sampled from a normal distribution with ($\mu = 0, \sigma = 1$). To maintain efficiency, 5 scenarios were sampled from the set of scenarios provided to evaluate the best 5 prescriptors on 

The final, timestep reward function was simply a composite of only two of the five metrics signified by the competition, chlorine violation bounds and cost of control. The following composite reward  was used to train the surrogate and indirectly evolve the prescriptors: 

\begin{align}
    r(C, A) &= \frac{-10}{N} \sum_{i=1}^N \left[ \max(c_i - 0.4,\ 0) + 5 \cdot \max(0.2 - c_i,\ 0) \right] \nonumber \\
    &\quad + \frac{5}{N} \sum_{i=1}^N 1_{\{c_i \in [0.2,\ 0.4]\}}\nonumber \\
    &\quad - 0.1 \cdot \sum_{j=1}^M a_j \nonumber \\
    &= \text{penalty}_{\text{Cl}} + \text{bonus}_{\text{Cl, in-range}} + \text{cost}_{\text{Cl}}, \nonumber \label{eq:3}
\end{align}

\noindent where $c_i$ are chlorine concentrations at each node $i$ and $a_j$ is the chlorine injection action made by the agent at injection node $j$. We found that, at least initially, lower bound chlorine violations dominated the chlorine penalty, so we weighed it more. To ensure that the agent was able to inject chlorine within a significant range, we also underemphasize the cost control objective.

\section{Results}
As was mentioned before, we initially only focused on two objectives to optimize, chlorine bound violations and cost of control. As a result, we only focus on limited results for said objectives. Analysis of the other 3 objectives showed at most little improvement throughout ESP-based evolution. ESP itself was ran for 4 iterations. 

Based on the final evaluation script and corresponding function, the final solutions had cost control values ranging from $[864, 3709]$. On the other hand, chlorine violation bounds were relatively similar in terms of range $[0.173, 0.175]$, suggesting little variance within populations and/or little, if any learning. 

\subsection{Pareto Front Analysis}
\begin{figure}[ht]
    \centering
    \includegraphics[width=0.75\linewidth]{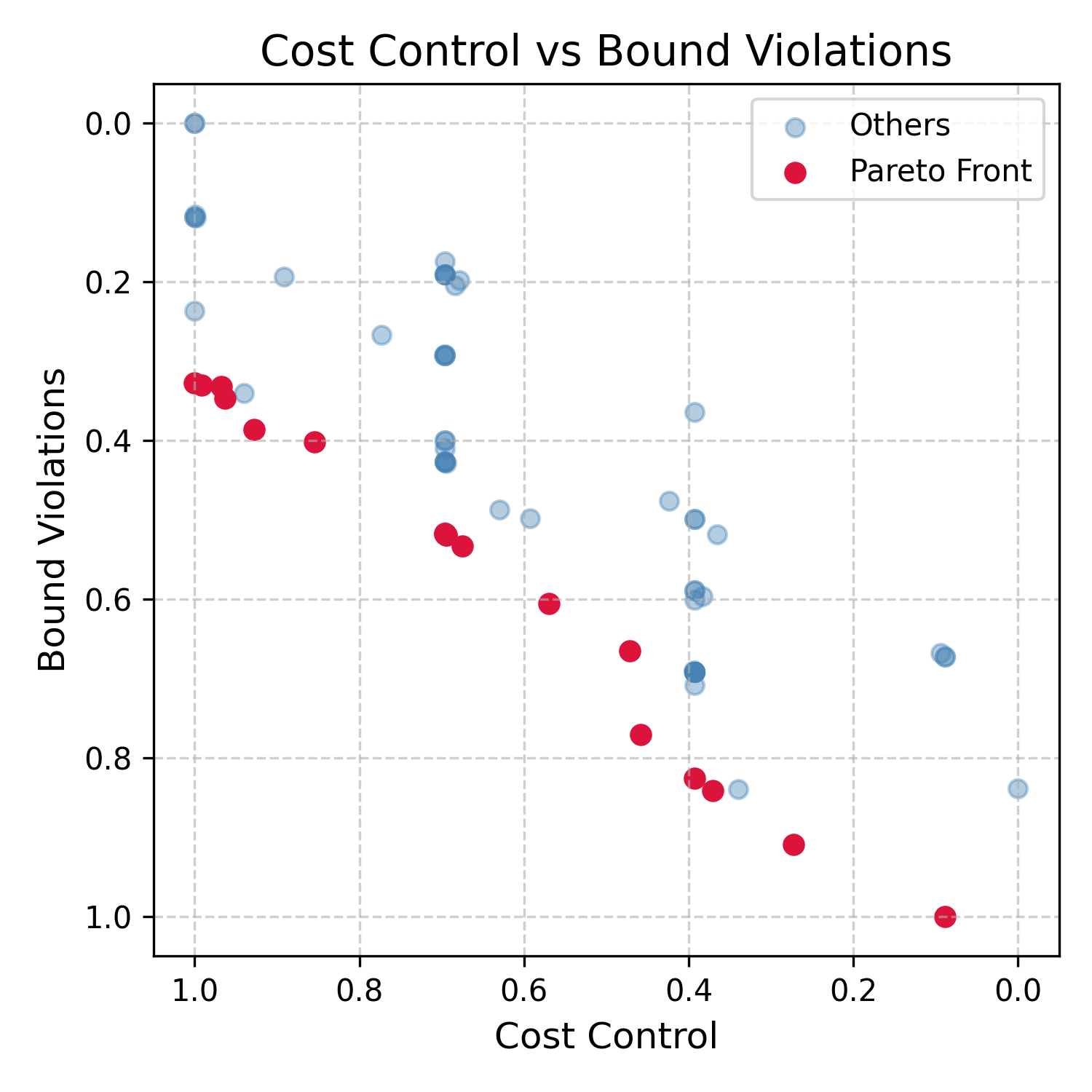}
    \caption{A 2D pareto front of the 60 best prescriptors generated from the ESP method after iteration 4. The values of each objective are normalized against the set of genomes. The pareto front is calculated and highlighted in the red.}
    \label{fig:1}
\end{figure}

To analyze the multi-objective nature of the task, we generate a Pareto front of the 60 best solutions from the terminal generation and iteration of the ESP cycle (iteration 4). These results are shown in [\ref{fig:1}]. Note that these values are normalized to emphasize the tradeoffs between each objective rather than absolute performance on the objectives. We can see an obvious linear front, indicating proportional trade-offs between both objectives. This is to be expected as more chlorine must be injected to keep chlorine concentration within bounds, especially in the early stages of learning where, in our case, lower bound violations dominated the violation penalty.

\section{Limitations and Future Work}
Of course, a major limitation of the current work is the inclusion of just 2 of the 5 objectives into our evolution scheme. To demonstrate holistic performance of the evolved agent, we intend to continue shaping the reward function to directly and indirectly optimize for each objective. In addition, utilizing other trained solutions to generate initial data for ESP will likely be necessary to widen the initial reward landscape and force learning. Lastly, implementation of curricular learning, by periodically changing the reward function by way of term coefficients and injection of data from good solutions would likely help the agents learn the system sequentially instead of being overwhelmed by the complex composite fitness. Ultimately, we would like to reach a point where the objectives can be weighed equally such that tradeoffs are relatively similar between each pair of objectives.

We plan on continuing this work as a proof of concept for ESP in real-world, complex agentic systems. 

\section{Participation Details}
A .zip file of the relevant source code was uploaded through the link provided. We will not be able to attend the IJCAI conference in-person.

\appendix

\section*{Ethical Statement}
There are no ethical issues.

\section*{Acknowledgments}
The configuration files for NEAT evolution as well as a general framework for surrogate architecture was created by Cognizant AI as part of the
COVID-19 XPRIZE Pandemic Response Challenge. The code used for NSGA-II was written by Hugo Aboud.

\bibliographystyle{named}
\bibliography{ijcai25}

\end{document}